\documentclass[10pt,twocolumn,letterpaper]{article}

\usepackage{wacv}              

\usepackage{graphicx}
\usepackage{amsmath}
\usepackage{amssymb}
\usepackage{booktabs}

\usepackage{algorithm, algpseudocode}
\usepackage{comment}
\usepackage{color}
\usepackage{booktabs}
\usepackage{multirow}
\usepackage{siunitx}
\usepackage[whole]{bxcjkjatype}
\usepackage{caption}
\input{mlpreamble}
\usepackage{bm}

\newcommand{\figref}[1]{Fig.~\ref{#1}}
\newcommand{\tabref}[1]{Tab.~\ref{#1}}
\newcommand{\equref}[1]{Eq.~(\ref{#1})}

\newcommand{\method}{ZeroPlantSeg\xspace}
\newcommand{\vsp}{-3mm}
\newcommand{\vspfig}{-5mm}

\definecolor{wacvblue}{rgb}{0.21,0.49,0.74}
\usepackage[pagebackref,breaklinks,colorlinks,allcolors=wacvblue]{hyperref}

\usepackage[capitalize]{cleveref}
\crefname{section}{Sec.}{Secs.}
\Crefname{section}{Section}{Sections}
\Crefname{table}{Table}{Tables}
\crefname{table}{Tab.}{Tabs.}

\def\wacvPaperID{280} 
\def\confName{WACV}
\def\confYear{2026}

\begin{document}

\title{Zero-shot Hierarchical Plant Segmentation \\via Foundation Segmentation Models and Text-to-image Attention}

\author{Junhao Xing$^{1}$
\qquad 
Ryohei Miyakawa$^{1}$
\qquad
Yang Yang$^{1}$
\qquad
Xinpeng Liu$^{1}$
\\
Risa Shinoda$^{1}$
\qquad
Hiroaki Santo$^{1}$
\qquad
Yosuke Toda$^{2,3}$
\qquad
Fumio Okura$^{1}$
\\
$^1$The University of Osaka \qquad $^2$Phytometrics \qquad $^3$Nagoya University
\\
{\tt\small \{xing.junhao, okura\}@ist.osaka-u.ac.jp} \qquad {\tt\small yosuke@phytometrics.jp}
}


\twocolumn[{%
\maketitle
        \centering
            \includegraphics[width=0.245\textwidth]{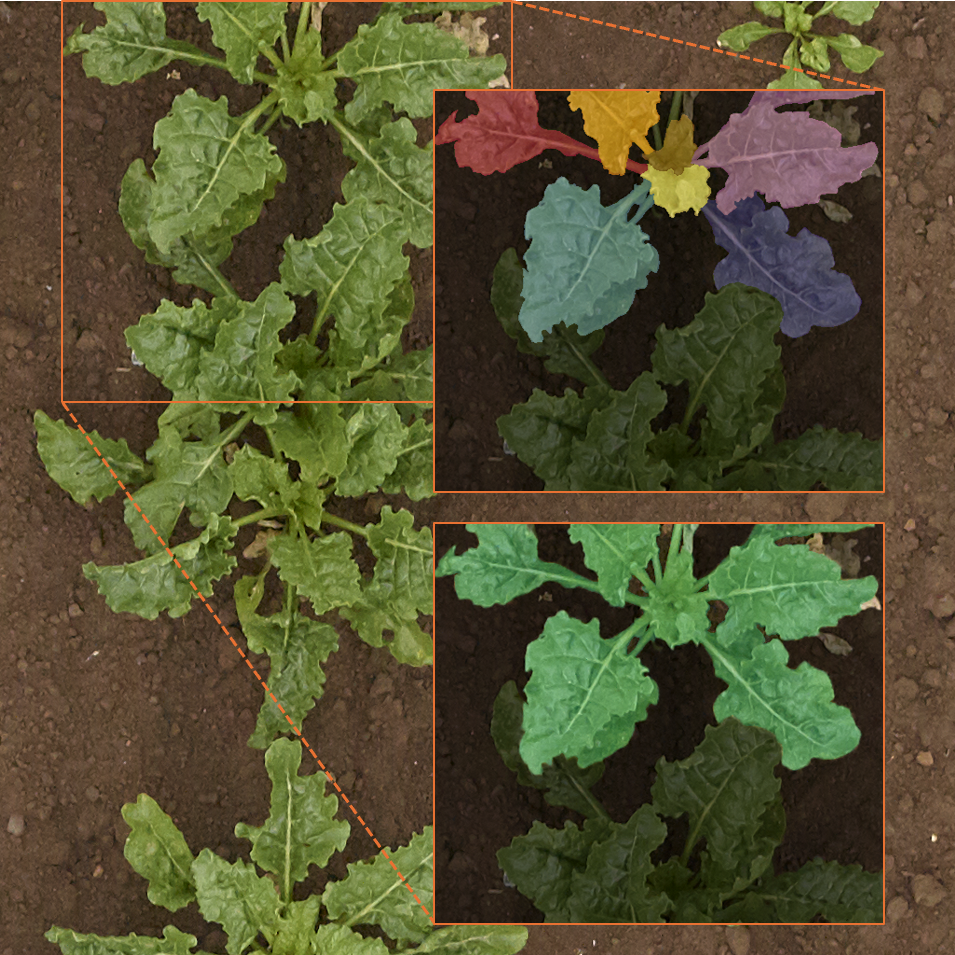} 
            \includegraphics[width=0.245\textwidth]{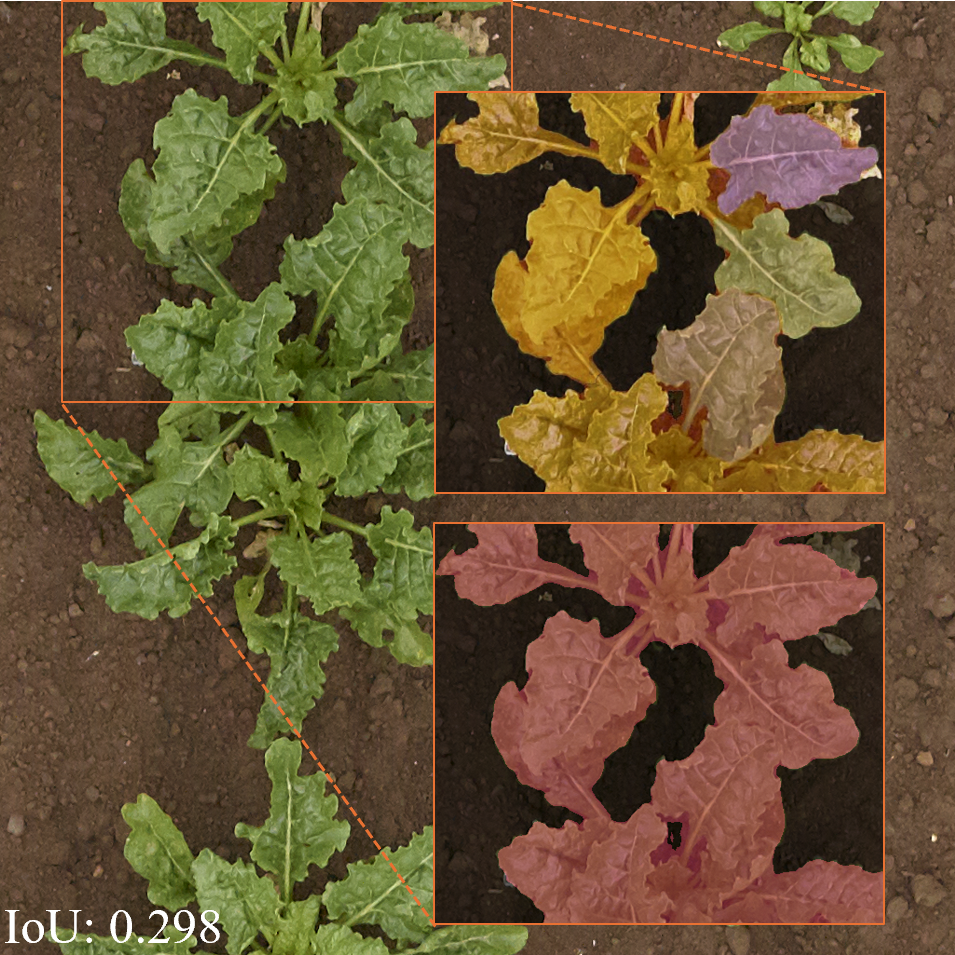}
            \includegraphics[width=0.245\textwidth]{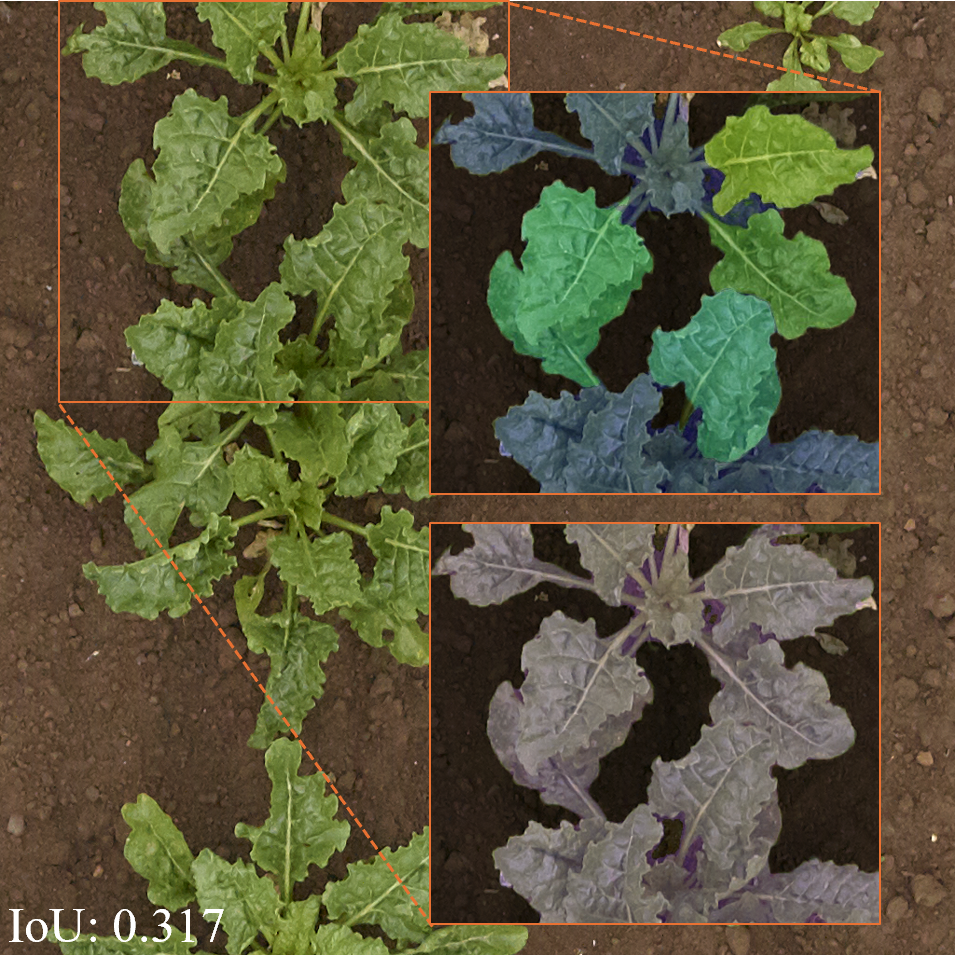}
            \includegraphics[width=0.245\textwidth]{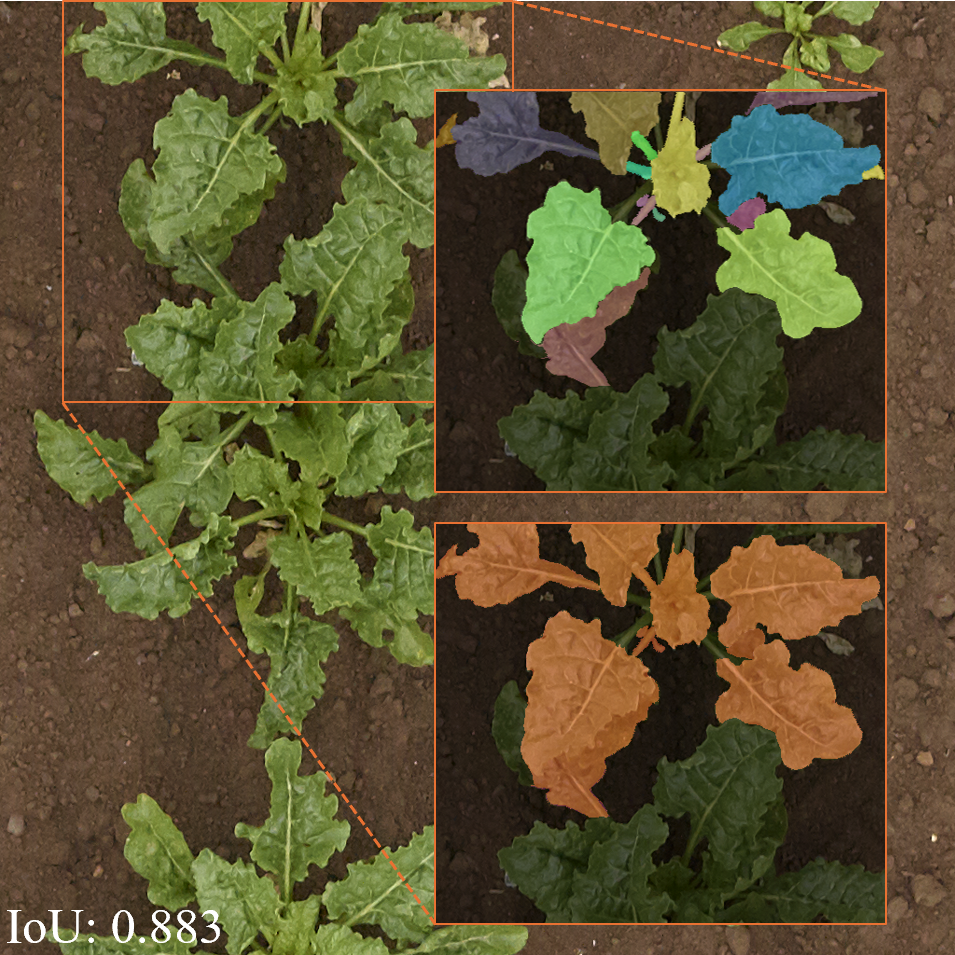}\\
            \makebox[0.245\linewidth]{\footnotesize (a) Ground-truth mask}
            \makebox[0.245\linewidth]{\footnotesize (b) G-SAM~\cite{GSAM}}
            \makebox[0.245\linewidth]{\footnotesize (c) G-SAM 2~\cite{ravi2024sam2segmentimages}}
            \makebox[0.245\linewidth]{\footnotesize (d) \method (Ours)}\vspace{-1mm}
        \captionof{figure}{\textbf{Zero-shot \emph{hierarchical} plant segmentation} by our \method in comparison with other zero-shot methods. Top: Leaf instance segmentation. Bottom: Plant individual segmentation. The IoU indicates the accuracy of plant individual segmentation.
        Our method achieves compelling results for both leaf instance and plant individual segmentation, validated by the high IoU score.
        \vspace{5mm}
        } \label{fig:result_sample}
}]

\begin{abstract}
    Foundation segmentation models achieve reasonable \textbf{leaf instance} extraction from top-view crop images without training (\ie, zero-shot). However, segmenting entire \textbf{plant individuals} with each consisting of multiple overlapping leaves remains challenging. This problem is referred to as a hierarchical segmentation task, typically requiring annotated training datasets that are often species-specific and require significant human labor. To address this, we introduce ZeroPlantSeg, a zero-shot segmentation for rosette-shaped plant individuals from top-view images. We integrate a foundation segmentation model, extracting leaf instances, and a vision-language model, reasoning about plants' structures to extract plant individuals without additional training. Evaluations on datasets with multiple plant species, growth stages, and shooting environments demonstrate that our method surpasses existing zero-shot methods and achieves better cross-domain performance than supervised methods. Implementations are available at \url{https://github.com/JunhaoXing/ZeroPlantSeg}.
\end{abstract}

    \section{Introduction}

    Computer vision is essential in image-based plant phenotyping, which extracts plants' traits from images for applications in plant sciences and agriculture.
    In particular, instance segmentation of leaves and plants is actively utilized for capturing important phenotypic attributes such as plant count, area, and morphology. 
    To this end, the joint segmentation of leaves and plant individuals is studied as \emph{hierarchical segmentation}~\cite{weylerJointPlantLeaf2022, weylerBaseline, HAPT}. Most existing methods rely on annotated training datasets, which are often species- or environment-specific. Thus, their practical usefulness is limited due to the required human labor.
    
    \textit{Zero-shot} segmentation models, such as the Segment Anything Model (SAM)~\cite{SAM}, offer a promising alternative by eliminating the need for domain-specific training. In agricultural applications, SAM and its variants have been actively utilized to segment \emph{leaf} instances (\eg, Leaf-only SAM~\cite{LSAM}), where several heuristics of plant leaves are introduced for accurate segmentation. However, a straightforward application of zero-shot segmentation models is not reasonable for segmenting \emph{plant individuals}, where their leaves have complex overlaps and mutual occlusions between individuals, as shown in \fref{fig:result_sample}(b).

    To address these issues, we propose \method, a zero-shot solution for hierarchical plant segmentation, targeting the top-view images capturing rosette-shaped plants, which is a common assumption in existing methods~\cite{weylerJointPlantLeaf2022, weylerBaseline, HAPT} and datasets~\cite{PhenoBench, Growliflower,SB20}. Similar to the existing zero-shot leaf segmentation methods (\eg, \cite{LSAM}), we use SAM and introduce plants' heuristics to extract leaf instances in the scene. To segment the plant individuals, we first leverage a prior from vision language models (VLMs), where we extract the \emph{stem} location from the leaf instances. Using the stem locations, we then perform a clustering-based method for combining the leaves belonging to the same individual.

    Experiments on three datasets (Phenobench~\cite{PhenoBench}, GrowliFlower~\cite{Growliflower}, and SB20~\cite{SB20}) show that our method outperforms a straightforward application of zero-shot segmentation. Besides, our method shows a notably better capability on cross-domain settings (\ie, different species and environments) compared to existing supervised methods for hierarchical segmentation~\cite{weylerJointPlantLeaf2022, weylerBaseline, HAPT}.

    \vspace{\vsp}
    \paragraph{Contributions}
    Our chief contributions are twofold:
    \begin{itemize}
        \setlength{\parskip}{0cm}
        \setlength{\itemsep}{0cm}
        \item We introduce the first zero-shot method for hierarchical plant segmentation, \method. Our pipeline addresses the challenge of segmenting overlapping plant structures without supervision by integrating a pre-trained SAM for leaf segmentation and a VLM-based attention for plant structure reasoning.
        \item Thanks to the zero-shot nature, \method performs robustly across domains (\eg, species) without retraining. Experiments show that our method outperforms zero-shot baselines and even supervised methods when applied to unseen plant species and environments.
    \end{itemize}

    \section{Related Work}
    Accurate segmentation of leaves and plants enables early identification of disease symptoms, such as spots, discoloration, or lesions on leaves. Also, it allows for precise measurement of leaf area, shape, and count, aiding in plant growth monitoring. However, classical segmentation methods relying on color-based~\cite{achanta2010slic} or brightness-based~\cite{Meyer1992ColorIS} thresholding struggle with the dynamic nature of plant growth, including variations in color, size, and structure. To overcome these limitations, learning-based approaches~\cite{he2017mask, romera2016recurrent} have been widely adopted, offering robustness against variations in lighting and plant morphology.
        
    \vspace{\vsp}
    \paragraph{Hierarchical plant segmentation}
    Hierarchical segmentation, \ie, simultaneous plant and leaf level segmentation, has recently garnered interest due to its ability to provide granular information, which indicates growth stages~\cite{PhenoBench}. Notable works~\cite{weylerJointPlantLeaf2022, weylerBaseline, HAPT} known as \emph{supervised} methods have explored hierarchical segmentation by leveraging plant structures rather than independently training separate models for leaves and plants. Furthermore, a \emph{self-supervised} method requiring fewer annotations~\cite{weylerPanopticSegmentationPartial2024} is proposed. However, these existing methods still require extensive annotated training data, incurring significant labor costs. Moreover, the generalizability of these models across plant species and diverse imaging conditions remains uncertain, as there are only limited datasets~\cite{CVPPPdataset, Growliflower, PhenoBench} for hierarchical segmentation.
    
    \vspace{\vsp}
    \paragraph{Zero-shot segmentation}
    \textit{Zero-shot} segmentation models offer a promising alternative by eliminating the need for domain-specific training~\cite{SEEM, Segnext}. Those models trained on vast datasets encompassing diverse scenarios with Vision Transformer (ViT)~\cite{ViT, SwinT} backbones can segment unseen images with remarkable generalization ability. For instance, SAM~\cite{SAM} is pre-trained on 11 million images containing over 1 billion masks (\ie, SA1B dataset~\cite{SAM}), enabling mask generation based on arbitrary object prompts. Furthermore, SAM provides the Automatic Mask Generator~(AMG) that produces masks for \textit{everything} in the image using a regular grid of point prompts. However, SAM's output is class-agnostic, making plant segmentation challenging in cluttered images. Methods integrating VLMs have emerged to introduce class awareness into segmentation outputs by learning adaptation for Contrastive Language-Image Pre-training (CLIP)~\cite{CLIP} or cross-attention~\cite{xu2015crossatt} between text and image features~\cite{OVSAM, GSAM, HIPIE}.

    In agricultural applications, SAM has been utilized to segment leaf instances from laboratory-controlled images and UAV-acquired field imagery. Leaf-only SAM proposed by Williams~\etal~\cite{LSAM} uses AMG on an image of potato plants and uses color, area, and shape filters to choose only leaf masks from class-agnostic masks. Torres-Lomas~\etal~\cite{TorresLomasSamForComprehensive} use AMG of SAM to segment grape berries from a bunch and measure their number and area. Yu~\etal~\cite{YuTimeSeriesField} capture images of a soybean canopy using a UAV, remove the background using a machine learning-based method, and segment individual leaves with SAM to perform variety classification based on leaf morphology. However, its performance degrades when applied to complex outdoor images, often merging multiple plants into a single mask. Fine-tuning SAM with object detection models or employing additional prompt-based refinements has been explored, but these approaches still require training data.

    \section{Method: \method}
    
    \begin{figure*}[tp]
        \centering
        \includegraphics[width=\linewidth]{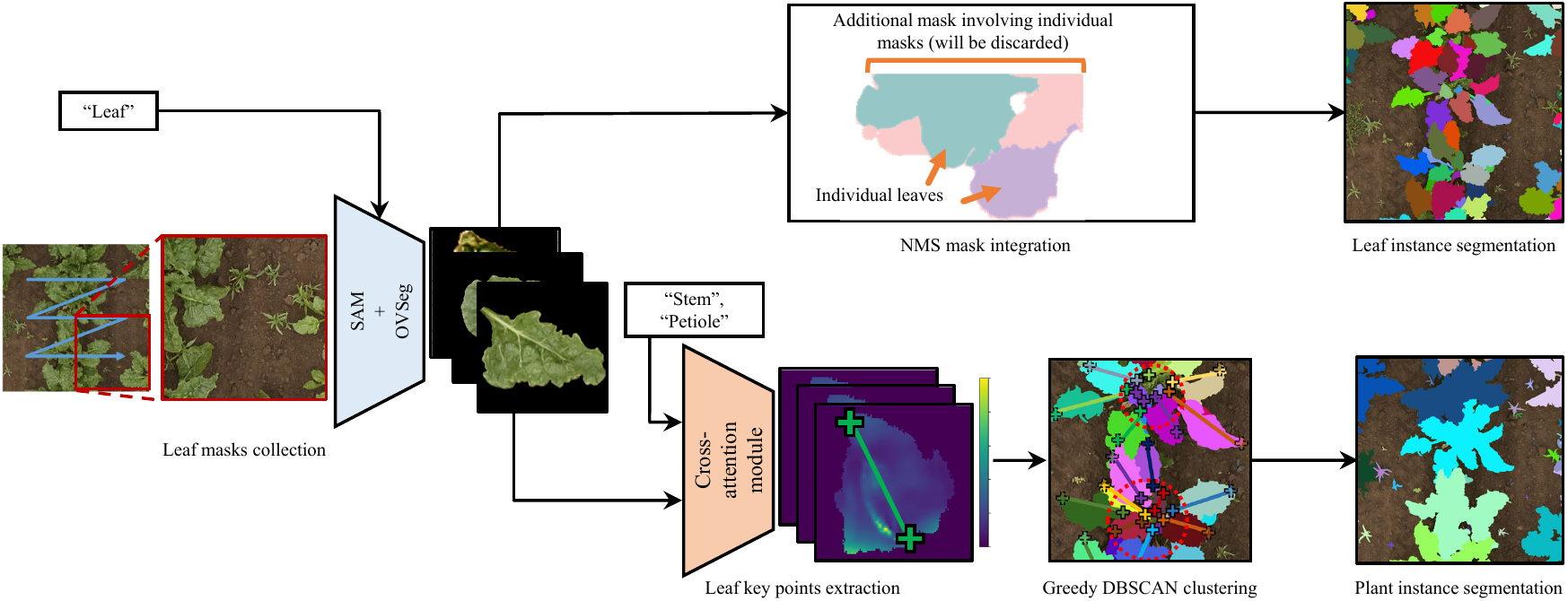}\vspace{-1mm}
        \caption{\textbf{Overview of \method.} An input image is sliced with a sliding window, and each patch is fed to the foundation segmentation model (\ie, SAM) to obtain all leaf masks. Those masks are integrated with an NMS to discard duplicates for leaf instance segmentation. For plant instance segmentation, the masks are input to the pre-trained cross-attention module to calculate their keypoints. Those points are used to estimate plant individual instances in our unsupervised greedy clustering.}
        \label{fig:framework}
    \end{figure*}
    
    \Fref{fig:framework} illustrates the overview of \method. We first input an RGB top-view image~$\bm{I}\in \mathbb{R}^{3\times H\times W}$ to the SAM-based foundation segmentation model, cropping with a sliding window to get the candidates of leaf regions. Since they have overlapping regions, we merge them using non-maximum suppression (NMS) to eliminate low-confidence masks and obtain leaf instance segmentation. On the other hand, we feed leaf candidate masks to the pre-trained cross-attention module of a VLM to identify the stem location and the base and tip points of each leaf. We finally cluster those points using an unsupervised clustering method to obtain plant instances as a union of leaf masks.
    
    \subsection{Leaf Instance Segmentation}
    As described below, we extract leaf instances using zero-shot segmentation models with text prompts.
    
    \vspace{\vsp}
    \paragraph{Leaf Candidates Extraction}
    We use the Automatic Mask Generator~(AMG) in SAM~\cite{SAM} to obtain the candidates of leaf masks from an input image. AMG generates masks from point prompts placed equally distributed on the image; however, obtaining masks for small instances compared to the image size is challenging, which is likely to occur in images captured from UAVs. We thus crop and upsample the input image using a sliding window to obtain masks for small leaves. Masks located at the window boundaries are not used. 

\begin{figure*}[tp]
        \centering
        \includegraphics[width=\linewidth]{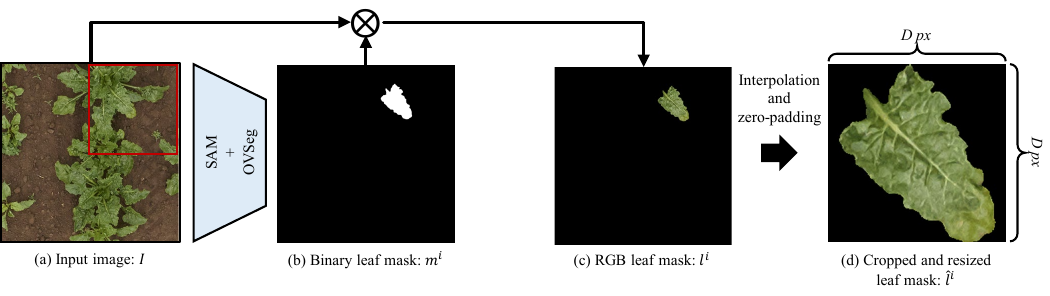} \vspace{-6mm}
        \caption{\textbf{The procedure to obtain leaf candidate masks and leaf images.} (a) Input RGB image. We crop the images with a sliding window (described as a red square in the figure) and enlarge them before being fed to a segmentation model for finer segmentation. (b) Binary leaf mask output by SAM and selected by OVSeg. (c) An RGB leaf image is obtained by multiplying the input image by the binary mask. (d) The image is cropped and resized to a square image of size~($D, D$). 
        }
        \label{fig:leaf_mask}
    \end{figure*}

    SAM does not classify the masks it produces, and its output contains masks for all class objects in the image. Therefore, we extract leaf masks by OVSeg~\cite{OVSeg}, a mask-adapted image classifier that takes the text and masked image as input. It encodes the mask and text prompt with a mask-adapted image encoder and text encoder, respectively, and calculates their cosine similarity to predict each mask's class. We use the text prompts ``green leaf'' and ``soil'' expecting that the scene only contains plants and soil. Masks with a larger similarity with ``soil'' than with ``green leaf'' are discarded. As shown in \figref{fig:leaf_mask}(b), we obtain binary masks of leaf, and we describe the set of those masks as~$\mathcal{M}=\{\bm{M}^{i}\}_{i=1}^{N}$, where~$\bm{M}^{i}\in \{0, 1\}^{H\times W}$.
    
    \vspace{\vsp}
    \paragraph{Mask Integration via NMS}
    The masks by AMG have overlapping regions, \ie, masks for two or more leaves are produced. Besides, since we use a sliding window to slice an image before inputting it to SAM, we have duplicated masks of the same instance in the overlapping region of the windows. To obtain unified leaf instances, we remove these duplicated masks. We use NMS in Leaf Only SAM~\cite{LSAM}, which combines or suppresses the overlapping masks. For details, please refer to~\cite{LSAM}. 

    \begin{figure}[tp]
        \centering
        \begin{subfigure}{\linewidth}
            \centering
            \includegraphics[width=\linewidth]{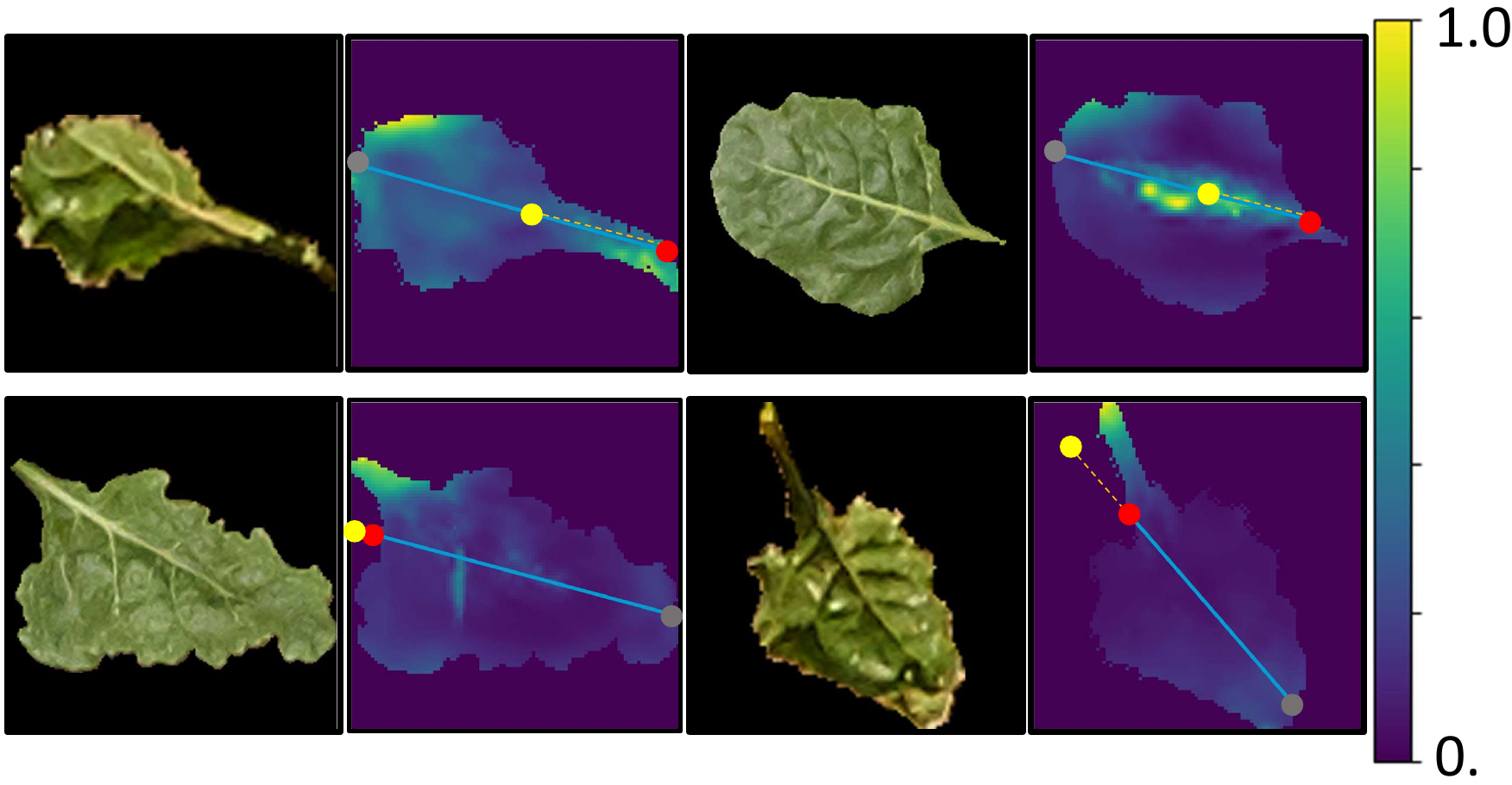}
            \caption{\label{fig:featmap} Leaf images and feature maps generated by the cross-attention with the text prompts ``stem'' and ``petiole''. Blue lines indicate the fitted WLS lines. Endpoints closer~(distances denoted as orange dotted lines) to the gravity centers~(yellow dots) are defined as base points~(red dots).}\vspace{2mm}
        \end{subfigure}
        \begin{subfigure}{\linewidth}
            \centering
            \includegraphics[width=\linewidth]{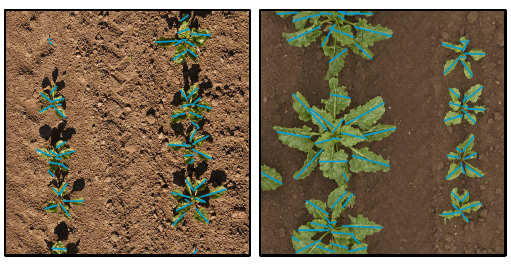}
            \caption{\label{fig:lines} Visualization of WLS lines. Their edge points are used to cluster leaf masks for plant individual segmentation.}
        \end{subfigure}
        \caption{The visualizations of feature maps and WLS lines (represented as blue lines in the figures) to obtain leaf keypoints.}
        \label{fig:featmap_line}
    \end{figure}

    \subsection{Plant Individual Segmentation}
    Our key contribution is the plant individual segmentation without supervision. Here, we describe the details of combining zero-shot segmentation and plants' heuristics.

    \vspace{\vsp}
    \paragraph{Keypoint Extraction via Text-to-Image Cross-Attention} 
    We segment plant individual instances by adequately grouping the obtained leaf masks. At the same time, it is challenging to extract plant instances correctly by simply clustering the masks when crops are densely planted, as shown in \figref{fig:result_sample}. Since the leaves of rosette-shaped plants grow from the center of the plants, a viable way is to extract the base points of the leaves from each leaf mask. However, the shape of the leaf blade is sometimes symmetric, and it can be ambiguous whether the apex-shaped point is the leaf's base or tip. Therefore, we here extract both the base and tip points of the leaves, hereafter called keypoints. 
    
    To obtain keypoints, we first prepare the leaf image set~$\mathcal{L}$, using binary masks~$\mathcal{M}$ as
    \begin{equation}
        \mathcal{L}=\{\bm{L}^{i} \in \mathbb{R}^{3\times H \times W} \mid \bm{L}^{i}=\bm{I}\otimes \bm{M}^{i}, \bm{M}^{i} \in \mathcal{M}\}_{i=1}^{N},
    \end{equation}
    where~$\otimes$ denotes pixel-wise multiplication with channel-wise broadcast (see~\fref{fig:leaf_mask}(c)). 
    We crop and resize each leaf image~$\bm{L}^{i}$ to a square image~$\hat{\bm{L}}^{i}$, as shown in~\figref{fig:leaf_mask}(d), obtaining a set of cropped leaf images as~\hbox{$\hat{\mathcal{L}}=\{\hat{\bm{L}}^{i} \in \mathbb{R}^{3\times D \times D}\}_{i=1} ^{N}$}.

    We then extract the \emph{stem} lines in the leaf images~$\hat{\mathcal{L}}$. We feed each~$\hat{\bm{L}}^{i}$ to the text-to-image cross-attention module of the pre-trained VLM with a text prompt that describes the stem-like part of a plant, namely, ``stem'' and ``petiole''. We use the output features of the text-to-image cross-attention (\ie, SwinT OGC) module of Grounding DINO~\cite{GDINO}. The features are spanned in four resolutions from~\hbox{$8\times8$} to~\hbox{$64\times64$}. We resize and average them to obtain the final feature map~$\bm{F}^{i}\in \mathbb{R}^{100\times 100}_{+}$, as shown in \fref{fig:featmap_line}(a).

    From the feature map, we estimate a stem line in the leaf image. We regress a straight line fitting to the features~$\bm{F}^{i}$ with weighted least squares (WLS) as indicated by the blue lines in \fref{fig:featmap_line}(a). We regard the line segment inside the leaf mask region as the stem, and the two endpoints of the line segment as the keypoints~\hbox{$\mathcal{Q}=\{q^{i}=(\bm{e}_{1}^{i}, \bm{e}_{2}^{i})\}_{i=1}^{N}$}. \Fref{fig:featmap_line}(b) visualizes the detected stems in the practical scenes. Supplementary materials describe technical details.

    \vspace{\vsp}
    \paragraph{Instance Grouping via Greedy Clustering}
    We use unsupervised clustering on the set of keypoints~$\mathcal{Q}$ to obtain plant instances. Although the most naive way is to apply a simple clustering method such as DBSCAN~\cite{DBSCAN}, it can produce invalid clusters since~$\mathcal{Q}$ includes both leaf tips and bases, where leaf tips are far from the plants' center. To address this issue, we perform the clustering in a greedy manner, assuming many leaf base points are located close to the center of the plants' individuals in the rosette-shaped plants.

    From all keypoints in~$q^{i}\in \mathcal{Q}$, we firstly define the ``petiole'', or the base keypoints of the leaves. We achieve it by calculating the feature maps' gravity centers and defining the closer endpoints as the base points. Then we select these base points~\hbox{$\hat{\mathcal{Q}}\subset\mathcal{Q}$} where a large number of points~(larger than a pre-defined constant~$n^*$) are located within a distance~$\epsilon$. We perform DBSCAN clustering for~$\hat{\mathcal{Q}}$ and select the largest cluster~$\mathcal{C}$, which is reliably considered to be a single plant's center. We then remove the leaf instances having the keypoints contained in~$\mathcal{C}$ as
    \begin{equation}
    \hat{\mathcal{Q}} \leftarrow \hat{\mathcal{Q}} - \theta_q\left(\theta_l(\mathcal{C})\right),        
    \end{equation}
    where~$\theta_l(\cdot)$ returns the set of leaf instances that contain keypoints in~$\mathcal{C}$, and~$\theta_q(\cdot)$ provides the keypoints in the corresponding leaf instances.    
    We repeat this process until no more clusters are formed or until the number of clusters reaches a pre-defined constant. We present the algorithm details in the supplementary materials.
    
    After the clustering, some points are regarded as outliers, not included in any cluster~\hbox{$\{\bm {o}^{k}\}_{k=1}^{N_o}\subset \hat{\mathcal{Q}}$}. Since we expect all detected leaf masks to be a part of a plant, we conduct post-processing that includes outliers in adequate clusters. Considering the clusters' shapes, we find the closest cluster to~$\bm{o}^{k}$ in terms of the Mahalanobis distance, as~\hbox{$c = \argmin_{c^*} d_{\text{MH}}(\bm{o}^k,\mathcal{C}^{c^*})$}, where~$c^*$ denotes a cluster ID. If~$d_{\text{MH}}$ is smaller than a threshold~$d_{th}$, we simply assign the cluster ID~$c$ to the leaf instance containing the keypoint~$\bm{o}^{k}$. If not, we consider that the leaf containing~$\bm{o}^k$ constructs a small plant instance with a single leaf, and we create a new cluster consisting of a single point~$\bm{o}^{k}$. Letting a new cluster as~$C'=\{\bm{o}^{k}\}$, the post process for~\hbox{$\{\bm {o}^{k}\}_{k=1}^{N_o}\subset \hat{\mathcal{Q}}$} is described as
    \begin{equation}
        \bm{o}^{k} \in \left\{
        \begin{array}{ll}
            C^c & \text{if}~d_{\text{MH}}(\bm{o}^k,\mathcal{C}^c) < d_{th} \\
            C'  & \text{otherwise}
        \end{array}
        \right.. \label{eq:postprocess}
    \end{equation}

    Finally, the plant individual instances are formed by merging leaf masks whose keypoints are included in the same cluster~$\mathcal{C}$. Therefore,~$j$-th plant instance~$\bm{P}^{j}$ is described as
    \begin{equation}
        \bm{P}^{j} = \bigcup_{\bm{M}\in \mathcal{M}^j}\bm{M},~\text{where}~\mathcal{M}^{j} = \{\bm{M}^{i} \mid \forall i~\mathrm{s.t.}~
        \bm{e}^{i} \in C^{j}\}. \label{eq:plant_ins}
    \end{equation}
    Additionally, if a single pixel in the resulting mask is potentially included in multiple plants, we do \textit{majority voting} to determine the belonging plant.

\section{Experiments}
    We compare our \method with supervised hierarchical segmentation by Weyler~\etal~\cite{weylerBaseline} and Gianmarco~\etal~\cite{HAPT}, which we hereafter call Weyler~\etal and HAPT, respectively. Besides, we also assess  \textit{zero-shot} segmentation models that do not require training data like our method, including LeafOnlySAM~\cite{LSAM}, Grounded SAM~\cite{GSAM}, and Grounded SAM 2~\cite{ravi2024sam2segmentimages}.

    \subsection{Dataset}
    \paragraph{PhenoBench~\cite{PhenoBench}} This is an RGB image dataset of sugar beet fields captured from a UAV.
    ~As the provided test data does not include annotations, we randomly divided the provided validation data and used half for validation and the other half for testing, resulting in $1407$ training, $386$ validation, and $386$ test images. 

    \vspace{\vsp}
    \paragraph{GrowliFlower~\cite{Growliflower}} This dataset contains RGB images of a cauliflower field located in Western Germany captured from a UAV. 
    ~We use images captured on August 12 and August 19, where the number of plants in each image is relatively large. We use $254$ images for validation, and $261$ for testing in total.
   
    \vspace{\vsp}
   \paragraph{SB20~\cite{SB20}} This is a dataset of sugar beet fields taken at Campus Klein Altendorf in Bonn, Germany, in 2020, using an agricultural field robot. 
   ~We randomly separate the annotated data: $50\%$  for validation and $50\%$ for testing, containing $72$ images each.
   
   \subsection{Implementation Details}
    For the SAM model, we use the ViT-H SAM with a \texttt{granularity} value of $0.8$. We set the leaf instance threshold~\hbox{$r_{th}=0.8$} for all experiments.
    To create a cropped leaf image~$\hat{\bm{L}}^i$, we set the image size as~\hbox{$800\times800$} pixels~(\ie, $D=800$), in which we use zero padding and bicubic interpolation for cropping and resizing.

   \subsection{Baselines}
   The setup for each baseline method is as follows. The settings of the hyperparameters of all models are carefully determined using the validation datasets.
   
   \vspace{\vsp}
   \paragraph{Supervised Models} We manually train the two supervised models on the \textbf{PhenoBench}~\cite{PhenoBench} dataset. We train \textbf{HAPT}~\cite{HAPT} with AdamW~\cite{AdamW} for~$200$ epochs with a batch size of~$8$. We set the step learning rate scheduler with an initial value of~\hbox{$4\times10^{-4}$} for backbone, and three exponential schedulers with initial learning rates of~\hbox{($4\times 10^{-4}$, $8\times10^{-4}$, and $8\times10^{-4}$)} for the semantic, plant instance, and leaf instance decoder, respectively. During training, we apply dropout with a probability of~$0.15$.
   We train \textbf{Weyler~\etal}~\cite{weylerBaseline} with Adam~\cite{Adam} for~$512$ epochs with a batch size of~$1$. We set the initial learning rate to~\hbox{$1\times10^{-3}$} and subsequently apply a polynomial learning rate decay~\hbox{$(1-\frac{e}{512})^{0.9}$}, where~$e$ is the current epoch.
   
    \vspace{\vsp}
   \paragraph{Zero-shot Models} We use the zero-shot models, including \textbf{Grounded SAM~(G-SAM)}~\cite{GSAM}, \textbf{Grounded SAM 2~(G-SAM2)}~\cite{ravi2024sam2segmentimages}, and \textbf{LeafOnlySAM~(LO-SAM)}~\cite{LSAM} with the pre-trained weights provided in their official repositories. 
   For the grounded version of SAMs (\ie, G-SAM~\cite{GSAM} and G-SAM2~\cite{ravi2024sam2segmentimages}), we use two text prompts, ``leaf'' and ``plant'', to obtain leaf and plant individual masks, respectively. 
   If a pixel is contained in multiple leaf masks, we assign the smallest one similarly to LO-SAM~\cite{LSAM}. 
   For plant individual masks, we assigned the largest mask for each pixel. 
   For leaf segmentation, predicted masks that cover more than 30~\% of the images are discarded. For plant segmentation, those that are larger than 50~\% are discarded. 
   LO-SAM~\cite{LSAM} is literally designed for segmenting \emph{leaf only}; thus, we do not assess the accuracy for the plant individual segmentation.

    \subsection{Metrics}
    We assess the standard metrics for segmentation accuracy: precision, recall, and average precision (AP). For all metrics, we use the Intersection over Union (IoU) threshold of $0.5$ and denote them as~$Prec_{50}$,~$Rec_{50}$, and~$AP_{50}$, respectively. Following previous research of hierarchical plant segmentation~\cite{PhenoBench}, we also evaluate the result with the Panoptic Quality (PQ)~\cite{PanopticSegmentation}, which assesses both segmentation quality and recognition quality. We compute~$PQ$ for plant instances and leaf instances, and denote them as~$PQ_{\text{plant}}$ and~$PQ_{\text{leaf}}$, respectively.

    \subsection{Results}
    We describe the experimental results for PhenoBench, GrowliFlower, and SB20 datasets. Since the supervised baselines (\ie, HAPT~\cite{HAPT} and Weyler~\etal~\cite{weylerBaseline}) are trained on the PhenoBench dataset, testing on GrowliFlower contains the domain gap between the training and testing datasets on plant species, and testing on the SB20 dataset contains the domain gap on the shooting environment. For each dataset, HAPT and Grounded SAM are selected as representative ones for supervised models and zero-shot models, respectively. For comparisons with other models~(\ie, G-SAM2~\cite{ravi2024sam2segmentimages} and LO-SAM~\cite{LSAM}), please refer to the supplementary materials.

    \begin{table}[tp]
        \centering
        \caption{Evaluation on the PhenoBench dataset. The best scores among the methods NOT trained on this dataset are marked \textbf{bold}.} \vspace{\vsp} 
        \resizebox{\hsize}{!}{
        \begin{tabular}{c|c|ccccc}
            \toprule                            & Trained on & \multicolumn{5}{c}{Metrics}  \\
             Approach                            & this dataset & $Prec_{50}~(\uparrow)$ & $Rec_{50}~(\uparrow)$ & $AP_{50}~(\uparrow)$ & $PQ_{\text{plant}}~(\uparrow)$ & $PQ_{\text{leaf}}~(\uparrow)$ \\
            \midrule HAPT~\cite{HAPT}                    & \checkmark & $66.75$           & $45.96$               & $37.40$         & $41.66$                   & $50.19$                  \\
            Weyler~\etal~\cite{weylerBaseline} & \checkmark & $44.26$                & $35.71$               & $24.63$              & $32.12$                        & $43.13$                       \\
            \midrule G-SAM~\cite{GSAM}                     & -          & $\bm{65.21}$                & $14.89$               & $13.80$              & $19.95$                        & $11.12$                       \\
            G-SAM2~\cite{ravi2024sam2segmentimages}& -          & $18.80$                & $30.08$               & $15.89$              & $18.27$                        & $10.19$                       \\
            LO-SAM~\cite{LSAM}                     & -          & $9.79$                & $13.50$               & $4.99$              & $ - $                        & $7.42$                       \\
            Ours                            & -          & $35.93$                & $\bm{46.09}$          & $\bm{25.32}$              & $\bm{32.21}$                        & $\bm{41.60}$                       \\
 
            \bottomrule
        \end{tabular}    
        }
        \label{tab:phenobench}
    \end{table}

    \begin{table}[tp]
        \centering
        \caption{Evaluation of each day by~$AP_{50}$ on the PhenoBench dataset~\cite{PhenoBench}.}\vspace{\vsp}
        \resizebox{\hsize}{!}{
        \begin{tabular}{c|c|ccc}
            \toprule                             & Trained on &   \multicolumn{3}{c}{$AP_{50}$}       \\
                  Approach                       & this dataset & May 15    & May 26    & June 5     \\
            \midrule HAPT~\cite{HAPT}                    & \checkmark & $47.84$ & $39.08$ & $22.55$      \\
            Weyler~\etal~\cite{weylerBaseline} & \checkmark & $31.84$      & $30.89$      & $14.65$      \\
             \midrule G-SAM~\cite{GSAM}                     & -          & $12.80$      & $18.68$      & $11.38$      \\
             G-SAM2~\cite{ravi2024sam2segmentimages}                     & -          & $14.12$      & $15.36$      & $20.98$      \\
             LO-SAM~\cite{LSAM}                     & -          & $2.74$      & $5.39$      & $7.44$      \\
            Ours                             & -          & $\bm{22.76}$      & $\bm{32.18}$      & $\bm{25.03}$ \\
            \bottomrule
        \end{tabular}
        }
        \label{tab:phenobench_dates}
    \end{table}

\begin{figure*}[tp]
        \centering
        \includegraphics[width=\linewidth]{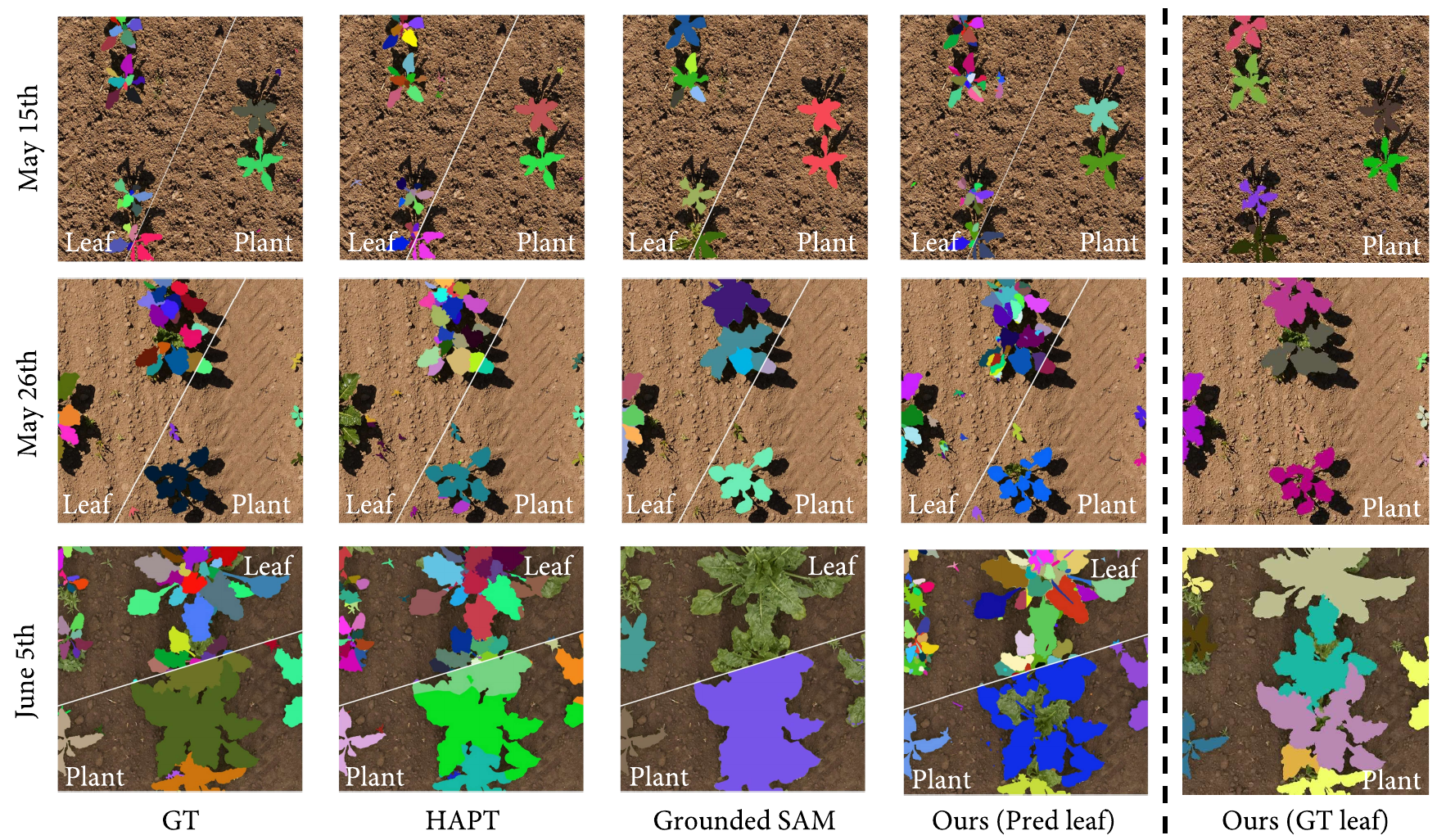}\vspace{-3mm}
        \caption{Segmentation results for leaves and plants on the PhenoBench dataset. Ablation study of using GT leaf instances is denoted as ``Ours (GT leaf)''. 
        }
        \label{fig:vis_phenobench}
    \end{figure*}

    \vspace{\vsp}
    \paragraph{PhenoBench---Same Plant Species and Shooting Environments}
    \Fref{fig:vis_phenobench} shows the visual results on the PhenoBench dataset, and \tabref{tab:phenobench} shows the quantitative evaluation. Our method with predicted leaf masks outperforms zero-shot models without any training data required, and achieves comparable accuracies to a supervised method (\ie, Weyler \etal~\cite{weylerBaseline}).
    
    For further analysis, an evaluation of the results by~$AP_{50}$ for each growth stage of the crop (date of dataset captured) is shown in \tabref{tab:phenobench_dates}. Overall, our method consistently achieves better accuracies than the unsupervised method (\ie, G-SAM~\cite{GSAM}). 
    Interestingly, our method achieves the best result on June 5, which is the most challenging case that contains complex overlaps among leaves (see \fref{fig:vis_phenobench}).

    \begin{table}[tp]
        \centering
        \caption{Evaluation on the GrowliFlower dataset containing the domain gap beyond plant species.}
    \vspace{\vsp}
        \resizebox{\hsize}{!}{
        \begin{tabular}{c|c|ccccc}
            \toprule                            & Trained on &  \multicolumn{5}{c}{Metrics} \\
             Approach                            & this dataset & $Prec_{50}~(\uparrow)$ & $Rec_{50}~(\uparrow)$ & $AP_{50}~(\uparrow)$ & $PQ_{\text{plant}}~(\uparrow)$ & $PQ_{\text{leaf}}~(\uparrow)$ \\
            \midrule HAPT \cite{HAPT}             &   -    & $6.45$                 & $3.93$                & $14.60$              & $15.70$                        & $2.57$                        \\
            Weyler~\etal~\cite{weylerBaseline}    &   -   & $0.06$                 & $0.12$                & $0.10$               & $0.03$                         & $0.00$                        \\
            G-SAM \cite{GSAM}              &   -    & $80.97$                & $64.06$               & $54.9$2              & $48.90$                        & $26.65$                       \\
            G-SAM2~\cite{ravi2024sam2segmentimages}& -          & $25.00$                & $16.67$               & $18.18$              & $20.06$                        & $15.38$                       \\
            LO-SAM~\cite{LSAM}                     & -          & $2.64$                & $2.38$               & $0.84$              & -                        & $1.47$                       \\
            Ours                      &   -   & $\bm{86.67}$           & $\bm{67.77}$          & $\bm{61.37}$         & $\bm{57.63}$                   & $\bm{36.63}$                  \\

            \bottomrule
        \end{tabular}
        }
        \label{tab:growliflower}
    \end{table}

    \begin{figure*}[tp]
        \centering
        \includegraphics[width=\linewidth]{
            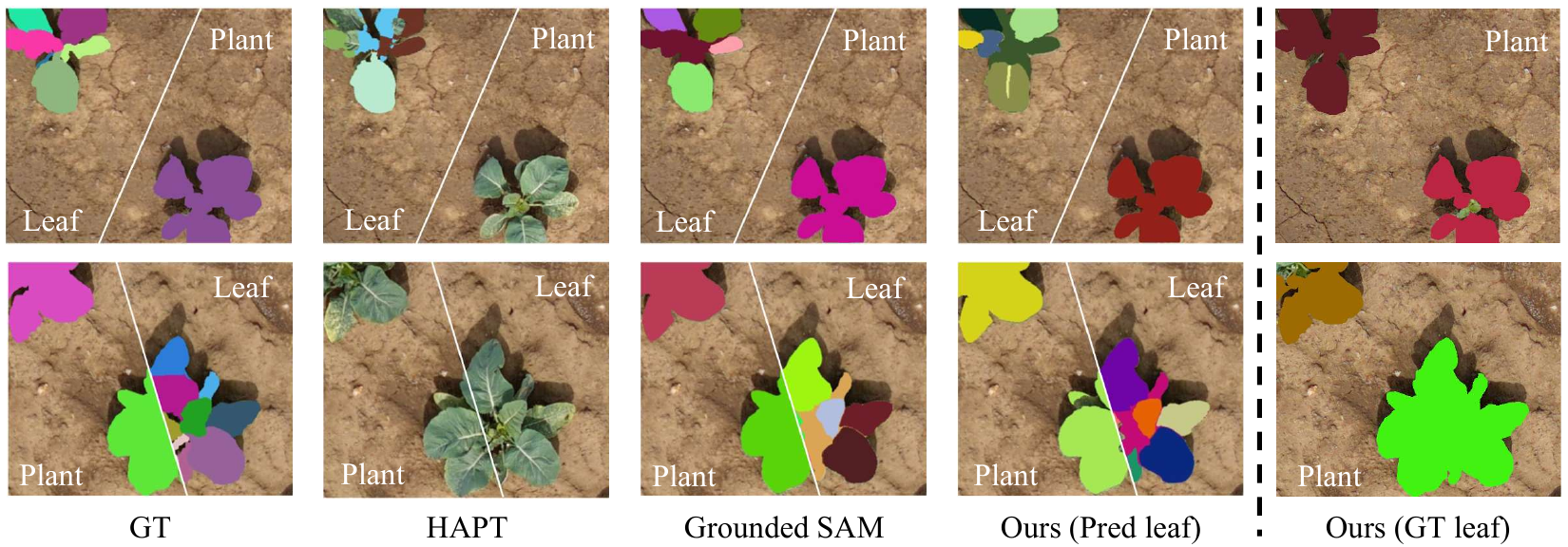
        }    \vspace{-7mm}
        \caption{Segmentation results for leaves and plants on the GrowliFlower dataset. Ablation study of using GT leaf instances is denoted as ``Ours (GT leaf)''. The supervised method ({HAPT}) is trained on the PhenoBench dataset, which exhibits a domain gap in plant species.}
        \label{fig:vis_growliflower}
    \end{figure*}

    \begin{table}[tp]
        \centering
        \caption{Evaluation on the SB20 dataset containing the domain gap beyond shooting environments.}
    \vspace{\vsp}
        \resizebox{\hsize}{!}{
        \begin{tabular}{c|c|cccc}
            \toprule                            & Trained on & \multicolumn{4}{c}{Metrics}   \\
             Approach                            & this dataset & $Prec_{50}~(\uparrow)$ & $Rec_{50}~(\uparrow)$ & $AP_{50}~(\uparrow)$ & $PQ_{\text{plant}}~(\uparrow)$  \\
            \midrule HAPT \cite{HAPT}              & - & $53.24$                & $8.17$                & $7.62$               & $11.89$                        \\
            Weyler~\etal~\cite{weylerBaseline}     & - & $0.00$                 & $0.00$                & $0.00$               & $0.00$                         \\
            G-SAM~\cite{GSAM}               & - & $\bm{86.56}$           & $18.75$               & $18.22$              & $25.47$                        \\
            G-SAM2~\cite{ravi2024sam2segmentimages}& -          & $43.58$                & $29.29$               & $22.47$              & $24.79$                        \\
            Ours                                   & - & $56.70$                & $\bm{30.04}$          & $\bm{24.44}$         & $\bm{28.04}$                   \\
            \bottomrule
        \end{tabular}
        }
        \label{tab:SB20}
    \end{table}
    
    \begin{figure*}[tp]
        \centering
        \includegraphics[width=\linewidth]{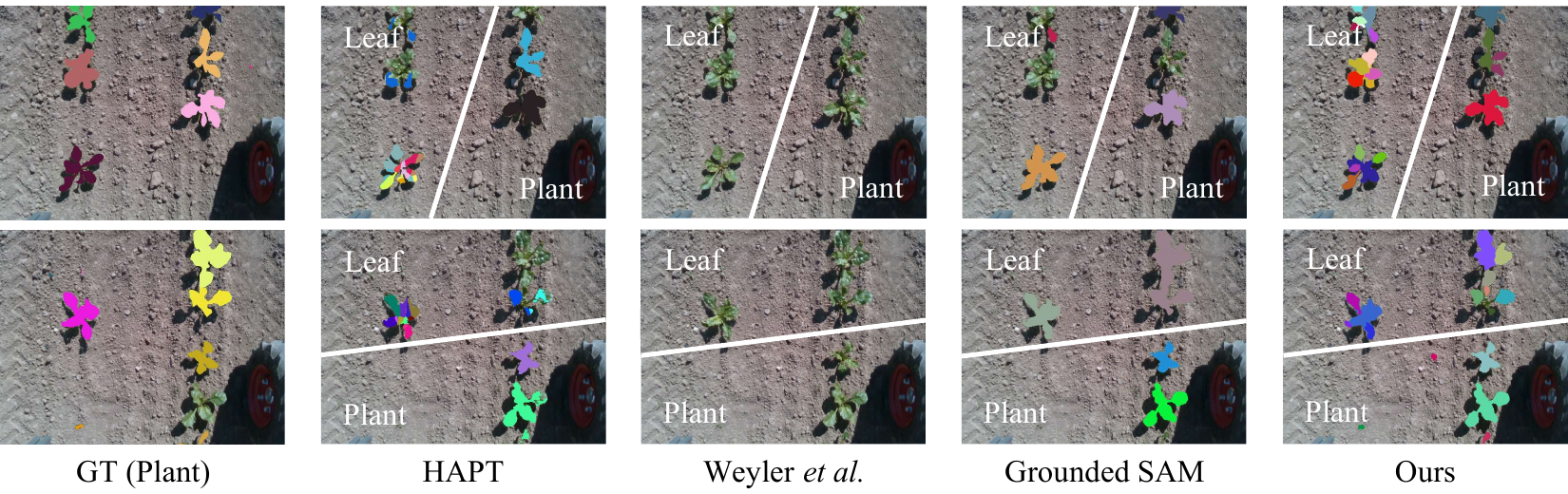}    \vspace{\vspfig}
        \caption{Visualization on the SB20 dataset, containing a domain gap of shooting environments for supervised methods.}
        \label{fig:vis_SB20}
    \end{figure*}

    \vspace{\vsp}
    \paragraph{GrowliFlower---Beyond Plant Species}
    \Tref{tab:growliflower} shows the evaluation of the test on the GrowliFlower dataset, cauliflower images recorded by UAV, where \figref{fig:vis_growliflower} shows the Visualization. For the supervised baselines trained with the PhenoBench dataset, this test dataset contains the domain gap between plant species~(\ie, sugar beet and cauliflower). Although HAPT and Weyler~\etal showed good performance on PhenoBench, which contains dense images with more occlusion than GrowliFlower, the score for the GrowliFlower dataset drops significantly. This result suggests that the current supervised methods do not generalize beyond the domain gap between training and testing data in plant species. On the other hand, our method outperforms all baselines in all metrics.

    \vspace{\vsp}
    \paragraph{SB20---Beyond Shooting Environments}
    SB20 is a dataset with different shooting environments~(\ie, UAV and ground vehicle, different farms) and the same plant species as PhenoBench.
    The accuracy and visualization of the test set on SB20 are shown in \tabref{tab:SB20} and \figref{fig:vis_SB20}, respectively. 
    Since SB20 only provides the ground-truth data for plant individual segmentation, we do not evaluate leaf instance segmentation accuracies, where LO-SAM is not involved in the experiment for this reason. 
    
    Compared with the \textit{zero-shot} baseline, \ie, Grounded SAM, our method yields higher scores in most of the metrics. HAPT shows a relatively high accuracy, but not as high as Grounded SAM and ours, and Weyler~\etal almost always fail to segment crops. These results indicate that the current state-of-the-art methods relying on training datasets~(\ie, HAPT and Weyler~\etal) do not generalize to the different shooting environments, even with the same plant species. 
    The domain gap of different shooting environments greatly influences the segmentation results for the supervised method, which implies the impracticability of current state-of-the-art methods.
    
    \subsection{Ablation Studies\label{sec:ablation}}
    We describe experiments to confirm the effectiveness of our key components: text-to-image attention for stem extraction and greedy clustering for plant instance grouping.

    \vspace{\vsp}
    \paragraph{Text-to-image Attention} We use text-to-image cross-attention to determine keypoints, including the leaves' bases. To evaluate the effectiveness of the cross-attention, we assess a straightforward alternative, where plant instances are grouped by clustering of the gravity centers of binary leaf masks instead of the base points. We use the same test dataset of PhenoBench~\cite{PhenoBench} as the main experiment.
    The result is summarized in the top rows in \tabref{tab:ablation}. Using text-to-image cross-attention outperforms the method that only uses the gravity centers of clusters. 
  
    \vspace{\vsp}  
    \paragraph{Greedy Clustering}
    To evaluate the effectiveness of the greedy clustering used for leaf grouping, we perform plant instance segmentation using the vanilla DBSCAN~\cite{DBSCAN} instead of our greedy one. The bottom rows in \tabref{tab:ablation} show that, although it is slightly inferior in~$Prec_{50}$, our algorithm outperforms the usual DBSCAN in all other metrics. 
    
    \vspace{\vsp}  
    \paragraph{Using the Ground-truth (GT) Leaf Instances}
    To exploit the theoretical maximum performance of our greedy clustering method, we conduct plant segmentation with the GT leaf instances. Visual results can be seen from~\fref{fig:vis_phenobench} and~\fref{fig:vis_growliflower} denoted as ``Ours (GT leaf)''. Our method using the GT leaf instances (see the bottom rows in~\tabref{tab:ablationleaf}) yields a notably high accuracy in~$Prec_{50}$ and~$PQ_{\text{plant}}$, indicating that our plant individual segmentation method leverages the further improvement of zero-shot segmentation models.
  
    \begin{table}[tp]
        \centering
        \caption{Ablation of text-to-image attention and greedy clustering.} 
    \vspace{\vsp}
        \resizebox{\hsize}{!}{
        \begin{tabular}{c|cccc}
            \toprule Text-to-image attention & $Prec_{50}~(\uparrow)$ & $Rec_{50}~(\uparrow)$ & $AP_{50}~(\uparrow)$ & $PQ_{\text{plant}}~(\uparrow)$ \\
            \midrule -               & $33.26$      & $44.62$      & $23.54$      & $29.97$             \\
            \checkmark               & $\bm{35.93}$ & $\bm{46.09}$ & $\bm{25.32}$ & $\bm{32.21}$        \\
            \bottomrule
            \toprule Greedy clustering & $Prec_{50}~(\uparrow)$ & $Rec_{50}~(\uparrow)$ & $AP_{50}~(\uparrow)$ & $PQ_{\text{plant}}~(\uparrow)$ \\
            \midrule -             & $\bm{36.83}$ & $43.73$      & $24.99$      & $31.24$             \\
            \checkmark             & $35.93$      & $\bm{46.09}$ & $\bm{25.32}$ & $\bm{32.21}$        \\
            \bottomrule
        \end{tabular}
        }
        \label{tab:ablation}
    \end{table}

    \begin{table}[tp]
        \centering
        \caption{Ablation study of using the GT leaf instances. We also list the results of our method using predicted leaf instances for reference. Using the GT instances indicates the upper-bound performance of our clustering-based plant individual segmentation.}
    \vspace{\vsp}
        \resizebox{\hsize}{!}{
        \begin{tabular}{c|c|cccc}
            \toprule Dataset & GT leaf instances & $Prec_{50}~(\uparrow)$ & $Rec_{50}~(\uparrow)$ & $AP_{50}~(\uparrow)$ & $PQ_{\text{plant}}~(\uparrow)$ \\
            \midrule 
            \multirow{2}{*}{PhenoBench} & -               & $35.93$                & ${46.09}$          & ${25.32}$              & ${32.21}$             \\
                                        &            \checkmark                & $\bm{84.79}$           & $\bm{53.64}$          & $\bm{48.93}$         & $\bm{65.77}$        \\
            \midrule
            \multirow{2}{*}{GrowliFlower}   & -  &  ${86.67}$           & $\bm{67.77}$          & ${61.37}$         & ${57.63}$ \\
                                            & \checkmark &  $\bm{95.80}$           & ${63.87}$               & $\bm{62.13}$         & $\bm{64.37}$ \\
            \bottomrule

        \end{tabular}
        }
        \label{tab:ablationleaf}
    \end{table}

    \section{Conclusions}
    We have proposed a zero-shot hierarchical segmentation framework, \method, for top-view imagery of rosette-shaped plants that leverages foundation segmentation models without requiring domain-specific training data. To overcome the limitations of overlapping plant boundaries, our method integrates a foundation segmentation model with a VLM and an unsupervised clustering to perform plant-level segmentation.
    Experimental results demonstrate that our method outperforms existing zero-shot segmentation baselines and state-of-the-art supervised methods under domain gaps in multiple plant species and agricultural fields. This suggests that our approach is a strong alternative to current supervised segmentation techniques in agricultural image analysis and plant phenotyping.

    \vspace{\vsp}
    \paragraph{Limitations} Although \method achieves a strong zero-shot capability for hierarchical segmentation, especially for plant individuals, the method is heavily based on the heuristics of rosette-shaped plants. While they are important in the agricultural domain, some major crops~(\eg, wheat) form different structures. Similar yet different heuristics for zero-shot segmentation for such crops should be introduced, which would be a viable future direction.
        
    \section*{Acknowledgments}
    \vspace{-1mm}
    This work was supported in part by the JSPS KAKENHI JP23H05491, JP25K03140, and JST FOREST JPMJFR206F.






\crefname{section}{Sec.}{Secs.}
\Crefname{section}{Section}{Sections}
\Crefname{table}{Table}{Tables}
\crefname{table}{Tab.}{Tabs.}

\renewcommand\thesection{\Alph{section}}
\renewcommand\thefigure{S\arabic{figure}}
\renewcommand\thetable{S\arabic{table}}
\renewcommand\theequation{S\arabic{equation}}



\begin{appendix}

\maketitlesupplementary 


This supplementary material provides additional information on technical details~(Secs.~\ref{supsec:featmap_keypoint} and \ref{supsec:clustering}). We also show additional visual results in \sref{supsec:visual}. 

\section{Technical Details}
 \subsection{Cross-Attention and Weighted Least Square Line\label{supsec:featmap_keypoint}} 
 In this section, we describe the text-to-image cross-attention that we use to obtain the center-like probability of each leaf mask and the weighted least square line that we use its edge points as key points of each leaf.

    \paragraph{Text-to-Image~Cross-Attention}
    Text-to-image cross-attention in one head is described as \equref{eq:crossattention}, where~\hbox{$\bm{F}_{I} \in \mathbb{R}^{N_I \times d}$} and~\hbox{$\bm{F}_{T} \in \mathbb{R}^{N_T \times d}$} denote the image feature and the text feature, respectively.
    \begin{equation}
        \text{CrossAttention}(\bm{F}_{I}, \bm{F}_{T}) = \text{softmax}\left(\frac{\bm{F}_{I}
        \bm{F}_{T}^{\mathsf{T}}}{\sqrt{d}}\right).~\label{eq:crossattention}
    \end{equation}
    In their official implementation \footnote{Grounding DINO, \url{https://github.com/IDEA-Research/GroundingDINO}, last accessed on July 10, 2025.}, the cross-attention module of Grounding DINO~\cite{GDINO} consists of six cross-attention layers. We use the output of the final layer to obtain the feature map. Specifically, we extract four feature maps in resolution from~$h/8 \times w/8$ to~$h/64 \times w/64$ from each head, where~$(h,w)$ denotes the size of the input image to the module. We resize to~$h/8 \times w /8$ and average them to obtain the feature map of the~$i$-th leaf mask. We describe it as~$f^{i} \in \mathbb{R}^{\frac{h}{8} \times \frac{w}{8}}_{+}$．
    
    \begin{algorithm}
        \caption{Greedy DBSCAN}
        \label{alg:greedyDBSCAN}
        \begin{algorithmic}
            [1] \Require{$\hat{\mathcal{Q}}$: The set of base points, $\epsilon$: The maximum distance to be regarded as a neighborhood in DBSCAN, $\text{InitMinPts}$: The initial minimum number of points to form a cluster in DBSCAN}
            \Ensure{$\mathcal{C}$: The set of clusters} \Function {GreedyDBSCAN}{$\hat{\mathcal{Q}}$, $\epsilon$, $\text{MinPts}$}
            \State $\mathcal{C}\gets \varnothing$ \State
            $\text{MinPts}\gets \text{InitMinPts}$ \For{$(j=1;j < num\_max\_cluster;j+=1)$}
            \State $\tilde{\mathcal{C}}\gets \mathrm{DBSCAN}(\hat{\mathcal{Q}},~\epsilon
            ,~\text{MinPts})$ \If {$\tilde{\mathcal{C}}==NULL~\textbf{and}~\text{MinPts}==2$}
            \State \textbf{break} \EndIf \State
            $C^{j} \gets \mathop{\arg \max}_{C \in \tilde{\mathcal{C}}}|C|$
            \State $\hat{\mathcal{Q}}\gets \hat{\mathcal{Q}}- C^{j}$ \State
            $\mathcal{C}\gets \mathcal{C}\cup C^{j}$ \If {$\text{MinPts}> 2$}
            \State $\text{MinPts}\gets \text{MinPts}- 1$ \EndIf \EndFor \State
            \Return $\mathcal{C}$ \EndFunction
        \end{algorithmic}
    \end{algorithm}
    
    \paragraph{Weighted Least Squares Line}
    Now, we have the feature map of the~$i$-th leaf mask,~$f^{i}$. To determine the key points of the mask, we calculate the weighted least squares (WLS) line on the feature map and obtain key points as its edge points. Considering the value at the coordinate~$(a, b)$ in the feature map~$f^{i}[a, b]$ as the weight at the point~$(a, b)$, an objective function of the WLS problem is described as \equref{eq:wls_object}, where~$\bm{x}=[x_{1}, x_{2}]^{\mathsf{T}}$ describes the WLS line.
    \begin{equation}
        \begin{gathered}
            J(\bm{x}) = ||\bm{W}^\frac{1}{2}(\bm{A}\bm{x}-\bm{b})||_2^2,\\ \text{where}~
            \bm{A} = \left[
            \begin{array}{cc}
                1      & a_1    \\
                1      & a_2    \\
                \vdots & \vdots \\
                1      & a_m    \\
            \end{array}\right], \bm{b} = \left[
            \begin{array}{c}
                b_1    \\
                b_2    \\
                \vdots \\
                b_m
            \end{array}\right],\\ \bm{W} = \left[
            \begin{matrix}
                f^{i}[a_{1}, b_{1}] &                    &        & \text{\large{0}}     \\
                                    & f^{i}[a_{2},b_{2}] &        &                     \\
                                    &                    & \ddots &                    &  \\
                \text{\large{0}}     &                    &        & f^{i}[a_{m}, b_{m}]
            \end{matrix}\right]. \label{eq:wls_object}
        \end{gathered}
    \end{equation}
    Solving \equref{eq:wls_object} with~$\frac{\partial J(\bm{x})}{\partial x_{1}}= \frac{\partial J(\bm{x})}{\partial x_{2}}= 0$, the WLS line~$\bm{x}$ can be calculated as~\equref{eq:wlsline}. For more detailed explanations, please refer to~\cite{wls}.
    \begin{equation}
        \bm{x}= (\bm{A}^{\mathsf{T}}\bm{W}\bm{A})^{-1}\bm{A}^{\mathsf{T}}\bm{W}\bm{b}.\label{eq:wlsline}
    \end{equation}

 \subsection{Greedy Clustering\label{supsec:clustering}} 
 Algorithm \ref{alg:greedyDBSCAN} shows the pseudo-code of the greedy clustering for plant individual segmentation.

\begin{figure*}[tp]
        \centering
        \includegraphics[width=\linewidth]{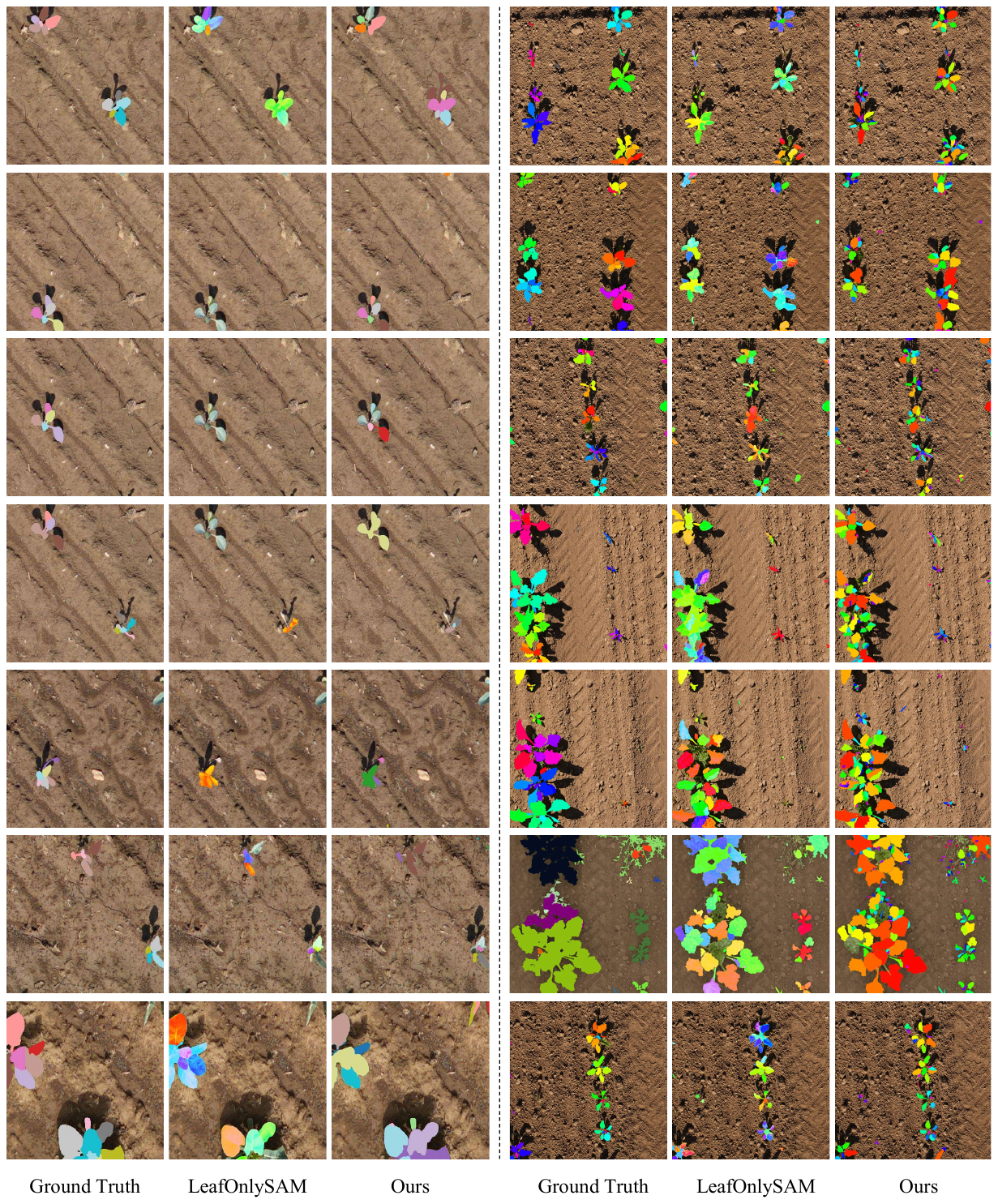}\vspace{-3mm}
        \caption{\textbf{Comparison of \method and LeafOnlySAM.} Since LeafOnlySAM does not involve plant segmentation, we only show the leaf instance segmentation results on the PhenoBench and Growliflower datasets. The left three columns show the results on the GrowliFlower dataset, and the right three columns show the results on the PhenoBench dataset.}
        \label{supfig:subvis1}
\end{figure*}

\begin{figure*}[tp]
        \centering
        \includegraphics[width=0.9\linewidth]{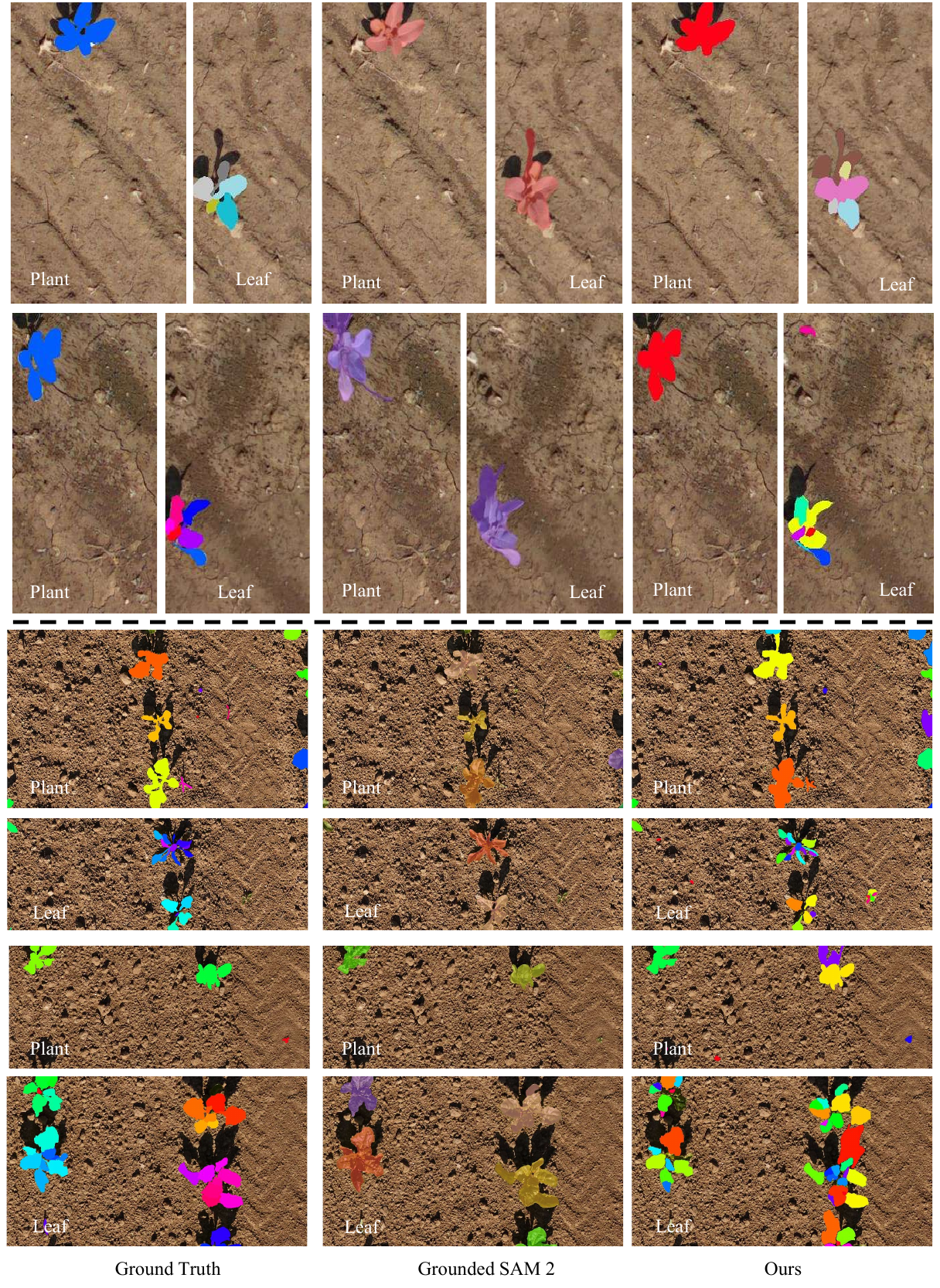}\vspace{-3mm}
        \caption{\textbf{Comparison of \method and Grounded SAM 2.} The results are divided into two parts. The top rows show the comparison on the GrowliFlower dataset, and the bottom rows show the comparison on the PhenoBench dataset. Each picture is denoted by ``Plant'' or ``Leaf'' to indicate whether it is a leaf instance segmentation or a plant individual segmentation result.}
        \label{supfig:subvis2}
\end{figure*}
\begin{figure*}[tp]
        \centering
        \includegraphics[width=0.9\linewidth]{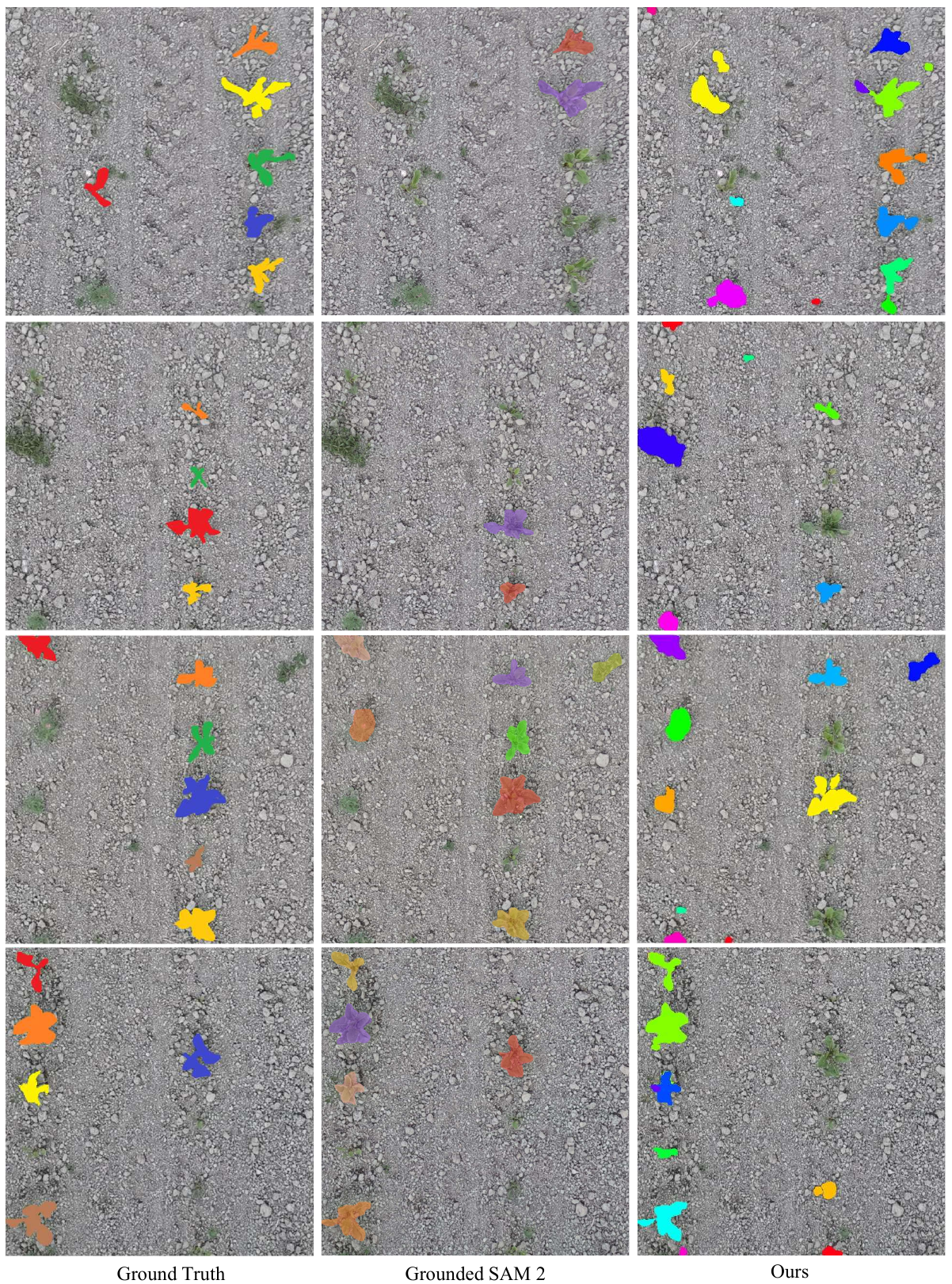}\vspace{-3mm}
        \caption{\textbf{Comparison of \method and Grounded SAM 2 on the SB20 dataset.} Since SB20 only supplies annotations of individual plants, we only visualize the plant segmentation results.}
        \label{supfig:subvis3}
\end{figure*}

\begin{figure*}[htp]
\centering
\includegraphics[width=0.9\linewidth]{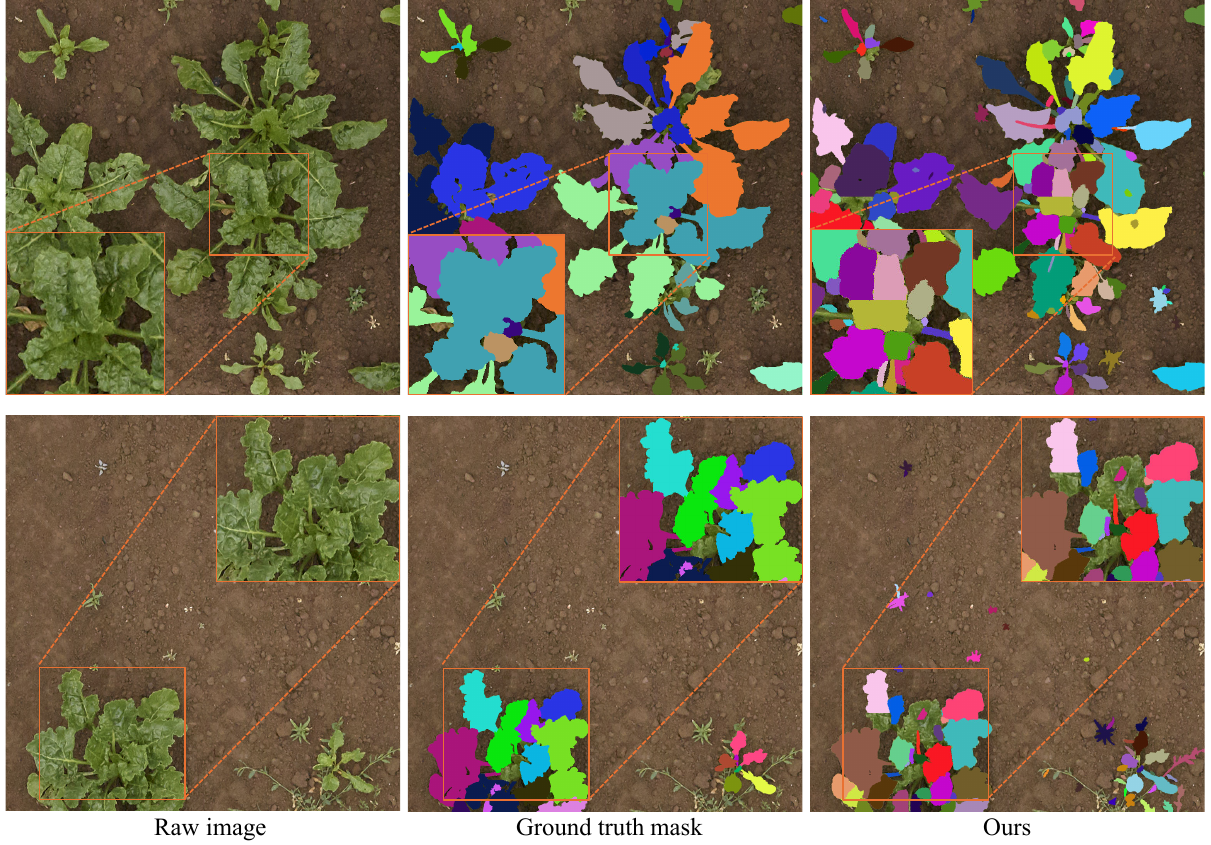}\vspace{-3mm}
\caption{\textbf{Failure cases of \method for leaf segmentation.}  The top row displays the incorrect split. The bottom row shows an example of failure to recognize leaves when they are occluded.}
\label{supfig:fail_leaf}

\includegraphics[width=0.9\linewidth]{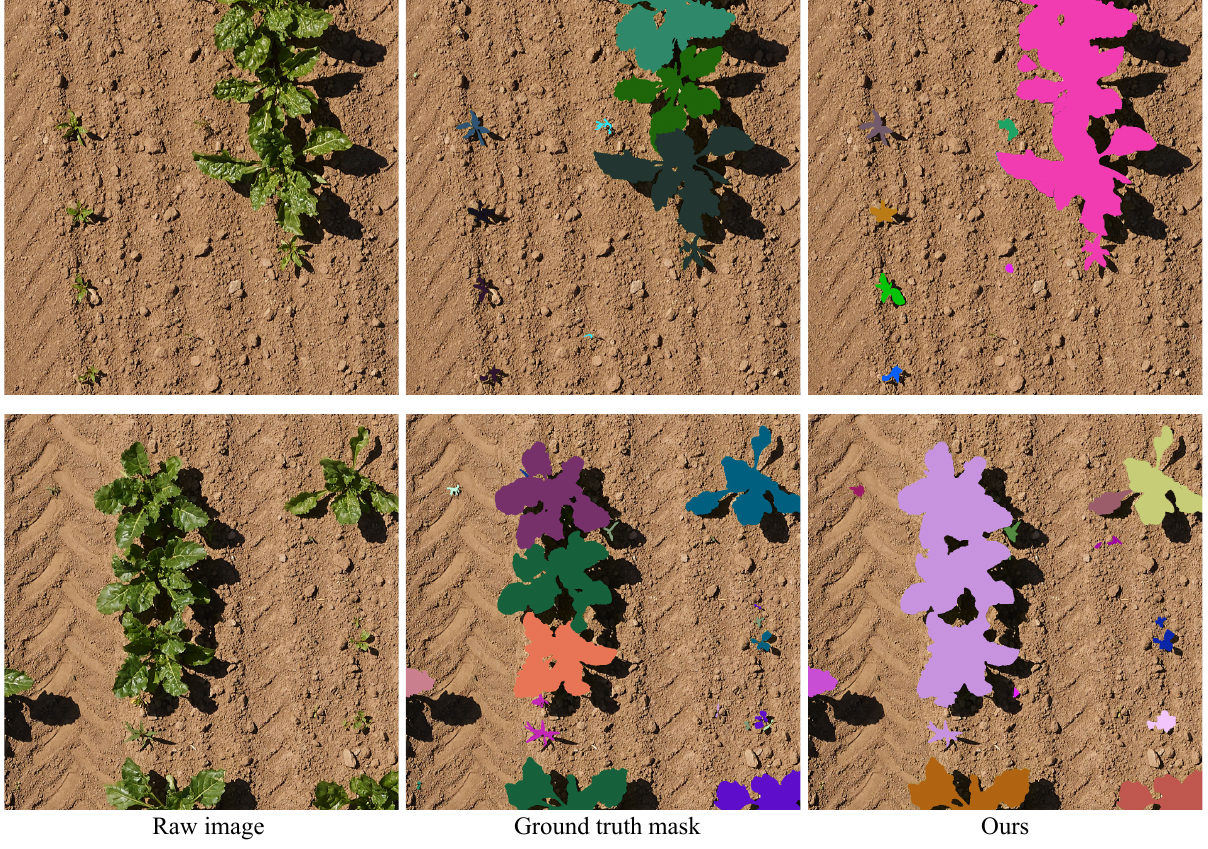}\vspace{-3mm}
\caption{\textbf{Failure cases of \method for plant segmentation. } These show the situation where leaves from different plants are mistakenly seen as a whole. }
\label{supfig:fail_plant}
\end{figure*}

 \section{Additional Results\label{supsec:visual}} 

While our method quantitatively surpasses these existing methods, Figs.~\ref{supfig:subvis1}, ~\ref{supfig:subvis2}, and \ref{supfig:subvis3} show the visual results for the recent zero-shot models, namely, Grounded SAM2~\cite{ravi2024sam2segmentimages} and LeafOnlySAM~\cite{LSAM}, compared to ours. Similar to the trend observed in the quantitative evaluation, our method more accurately estimates individual instances of leaves and plants than the state-of-the-art zero-shot segmentation models.

On the other hand, our method sometimes fails to recognize leaves where they are occluded by other leaves, or mistakenly splits one leaf into different parts (Fig.~\ref{supfig:fail_leaf}). This could be caused by the sliding windows strategy, as it would split one leaf into different windows and see them as individual leaves.
In the meantime, it might also lead to incorrect plant segmentation by mixing leaves as if they were from the same plant. This could be caused by aggressive thresholds when executing Greedy DBSCAN (Fig.~\ref{supfig:fail_plant}).

\end{appendix}

{\small
\bibliographystyle{ieee_fullname}
\bibliography{ref}
}

\end{document}